\documentclass[conference]{IEEEtran}

\usepackage{amsmath}
\usepackage{cite}

\usepackage{color}

\ifCLASSINFOpdf
   \usepackage[pdftex]{graphicx}
   \graphicspath{{../pdf/}{../jpeg/}}
   \DeclareGraphicsExtensions{.pdf,.jpeg,.png}
\else
   \usepackage[dvips]{graphicx}
   \graphicspath{{../eps/}}
   \DeclareGraphicsExtensions{.eps}
\fi

\usepackage{hyperref}

\usepackage{booktabs}
\usepackage{multirow}
\begin{document}
\title{EMNIST: an extension of MNIST to handwritten letters}

\author{\IEEEauthorblockN{Gregory Cohen, Saeed Afshar, Jonathan Tapson, and Andr\'e van Schaik }
\IEEEauthorblockA{The MARCS Institute for Brain, Behaviour and Development\\
Western Sydney University\\
Penrith, Australia 2751\\
Email: g.cohen@westernsydney.edu.au}}

\maketitle

\begin{abstract}
The MNIST dataset has become a standard benchmark for learning, classification and computer vision systems. Contributing to its widespread adoption are the understandable and intuitive nature of the task, its relatively small size and storage requirements and the accessibility and ease-of-use of the database itself. The MNIST database was derived from a larger dataset known as the NIST Special Database 19 which contains digits, uppercase and lowercase handwritten letters. This paper introduces a variant of the full NIST dataset, which we have called Extended MNIST (EMNIST), which follows the same conversion paradigm used to create the MNIST dataset. The result is a set of datasets that constitute a more challenging classification tasks involving letters and digits, and that shares the same image structure and parameters as the original MNIST task, allowing for direct compatibility with all existing classifiers and systems. Benchmark results are presented along with a validation of the conversion process through the comparison of the classification results on converted NIST digits and the MNIST digits.
\end{abstract}


\IEEEpeerreviewmaketitle

\section{Introduction}
The importance of good benchmarks and standardized problems cannot be understated, especially in competitive and fast-paced fields such as machine learning and computer vision. Such tasks provide a quick, quantitative and fair means of analyzing and comparing different learning approaches and techniques. This allows researchers to quickly gain insight into the performance and peculiarities of methods and algorithms, especially when the task is an intuitive and conceptually simple one. 

As single dataset may only cover a specific task, the existence of a varied suite of benchmark tasks is important in allowing a more holistic approach to assessing and characterizing the performance of an algorithm or system. In the machine learning community, there are several standardized datasets that are widely used and have become highly competitive. These include the MNIST dataset \cite{LeCun1998}, the CIFAR-10 and CIFAR-100 \cite{Krizhevsky2009} datasets,  the STL-10 dataset \cite{Coates2011}, and Street View House Numbers (SVHN) dataset \cite{Netzer2011}.

Comprising a 10-class handwritten digit classification task and first introduced in 1998, the MNIST dataset remains the most widely known and used dataset in the computer vision and neural networks community. However, a good dataset needs to represent a sufficiently challenging problem to make it both useful and to ensure its longevity \cite{Orchard2015b}. This is perhaps where MNIST has suffered in the face of the increasingly high accuracies achieved using deep learning and convolutional neural networks. Multiple research groups have published accuracies above $99.7\%$ \cite{Wan2013, Ciresan2012a, Sato2015, Chang2015, Lee2015a}, a classification accuracy at which the dataset labeling can be called into question. Thus, it has become more of a means to test and validate a classification system than a meaningful or challenging benchmark. 

The accessibility of the MNIST dataset has almost certainly contributed to its widespread use. The entire dataset is relatively small (by comparison to more recent benchmarking datasets), free to access and use, and is encoded and stored in an entirely straightforward manner. The encoding does not make use of complex storage structures, compression, or proprietary data formats. For this reason, it is remarkably easy to access and include the dataset from any platform or through any programming language. 

The MNIST database is a subset of a much larger dataset known as the NIST Special Database 19 \cite{Grother1995}. This dataset contains both handwritten numerals and letters and represents a much larger and more extensive classification task, along with the possibility of adding more complex tasks such as writer identification, transcription tasks and case detection.

The NIST dataset, by contrast to MNIST, has remained difficult to access and use. Driven by the higher cost and availability of storage when it was collected, the NIST dataset was originally stored in a remarkably efficient and compact manner. Although source code to access the data is provided, it remains challenging to use on modern computing platforms. For this reason, the NIST recently released a second edition of the NIST dataset \cite{Grother2016}. The second edition of the dataset is easier to access, but the structure of the dataset, and the images contained within, differ from that of MNIST and are not directly compatible.

The NIST dataset has been used occasionally in neural network systems. Many classifiers make use of only the digit classes \cite{Milgram2005, Granger2007}, whilst others tackle the letter classes as well \cite{Radtke2008, Ciresan2011, Koerich2005, Cavalin2006}. Each paper tackles the task of formulating the classification tasks in a slightly different manner, varying such fundamental aspects as the number of classes to include, the training and testing splits, and the preprocessing of the images. 

In order to bolster the use of this dataset, there is a clear need to create a suite of well-defined datasets that thoroughly specify the nature of the classification task and the structure of the dataset, thereby allowing for easy and direct comparisons between sets of results.

This paper introduces such a suite of datasets, known as Extended Modified NIST (EMNIST). Derived from the NIST Special Database 19, these datasets are intended to represent a more challenging classification task for neural networks and learning systems. By directly matching the image specifications, dataset organization and file formats found in the original MNIST dataset, these datasets are designed as drop-in replacements for existing networks and systems. 

This paper introduces these datasets, documents the conversion process used to create the images, and presents a set of benchmark results for the dataset. These results are then used to further characterize and validate the datasets.

The EMNIST dataset can be accessed and downloaded from \url{https://www.westernsydney.edu.au/bens/home/reproducible_research/emnist}.

\subsection{The MNIST and NIST Dataset}
\label{sec:the-datasets}

The NIST Special Database 19 \cite{Grother1995} contains handwritten digits and characters collected from over 500 writers. The dataset contains binary scans of the handwriting sample collection forms, and individually segmented and labeled characters which were extracted from the forms. The characters include numerical digits and both uppercase and lowercase letters. The database was published as a complete collection in 1995 \cite{Grother1995}, and then re-released using a more modern file format in September 2016 \cite{Grother2016}. The dataset itself contains, and supersedes, a number of previously released NIST handwriting datasets, such as the Special Databases 1, 3 and 7. 

The MNIST dataset is derived from a small subset of the numerical digits contained within the NIST Special Databases 1 and 3, and were converted using the method outlined in \cite{LeCun1998}. The NIST Special Database 19, which represents the final collection of handwritten characters in that series of datasets, contains additional handwritten digits and an extensive collection of uppercase and lowercase handwritten letters. 

The authors and collators of the NIST dataset also suggest that the data contained in Special Database 7 (which is included in Special Database 19) be used exclusively as a testing set as the samples were collected from high school students and pose a more challenging problem. 

The NIST dataset was intended to provide multiple optical character recognition tasks and therefore presents the character data in five separate organizations, referred to as data hierarchies. There are as follows:

\begin{itemize}
\item \textbf{\textit{By\_Page}}: This hierarchy contains the unprocessed full-page binary scans of handwriting sample forms. The character data used in the other hierarchies was collected through a standardized set of forms which the writers were asked to complete. 3699 forms were completed.
\item \textbf{\textit{By\_Author}}: This hierarchy contains individually segmented handwritten characters images organized by writer. It allows for such tasks as writer identification but offers little in the way of classification benefit as each grouping contains digits from multiple classes.
\item \textbf{\textit{By\_Field}}: This organization contains the digits and character sorted by the field on the collection form in which they appear. This is primarily useful for segmenting the digit classes as they appear in their own isolated fields.
\item \textbf{\textit{By\_Class}}: This represents the most useful organization from a classification perspective as it contains the segmented digits and characters arranged by class. There are 62 classes comprising [0-9], [a-z] and [A-Z]. The data is also split into a suggested training and testing set. 
\item \textbf{\textit{By\_Merge}}: This data hierarchy addresses an interesting problem in the classification of handwritten digits, which is the similarity between certain uppercase and lowercase letters. Indeed, these effects are often plainly visible when examining the confusion matrix resulting from the full classification task on the \textit{By\_Class} dataset. This variant on the dataset merges certain classes, creating a 47-class classification task. The merged classes, as suggested by the NIST, are for the letters C, I, J, K, L, M, O, P, S, U, V, W, X, Y and Z.
\end{itemize}

The conversion process described in this paper and the provided code is applicable to all hierarchies with the exception of the \textit{By\_Page} hierarchy as it contains fundamentally different images. However, the primary focus of this work rests with the \textit{By\_Class} and \textit{By\_Merge} organizations as they encompass classification tasks that are directly compatible with the standard MNIST dataset classification task. 

\begin{table}[]
\centering
\caption{Breakdown of the number of available training and testing samples in the NIST Special Database 19 using the original training and testing splits.}
\label{tab:dataset-composition}
\begin{tabular}{@{}llcccc@{}}
\toprule
 & Type & \multicolumn{1}{l}{No. Classes} & \multicolumn{1}{l}{Training} & \multicolumn{1}{l}{Testing} & \multicolumn{1}{l}{Total} \\ \midrule
By Class & Digits & 10 & 344,307 & 58,646 & 402,953 \\
 & Uppercase & 26 & 208,363 & 11,941 & 220,304 \\
 & Lowercase & 26 & 178,998 & 12,000 & 190,998 \\
 & \textbf{Total} & \textbf{62} & \textbf{731,668} & \textbf{82,587} & \textbf{814,255} \\
 &  & \multicolumn{1}{l}{} & \multicolumn{1}{l}{} & \multicolumn{1}{l}{} & \multicolumn{1}{l}{} \\
By Merge & Digits & 10 & 344,307 & 58,646 & 402,953 \\
 & Letters & 37 & 387,361 & 23,941 & 411,302 \\
 & \textbf{Total} & \textbf{47} & \textbf{731,668} & \textbf{82,587} & \textbf{814,255} \\
 &  & \multicolumn{1}{l}{} & \multicolumn{1}{l}{} & \multicolumn{1}{l}{} & \multicolumn{1}{l}{} \\
MNIST \cite{LeCun1998} & Digits & 10 & 60,000 & 10,000 & 70,000 \\ \bottomrule
\end{tabular}
\end{table}

Table~\ref{tab:dataset-composition} shows the breakdown of the original training and testing sets specified in the releases of the NIST Special Database 19. Both the \textit{By\_Class} and \textit{By\_Merge} hierarchies contain 814,255 handwritten characters consisting of a suggested 731,668 training samples and 82,587 testing samples. It should be noted however, that almost half of the total samples are handwritten digits. 

The \textit{By\_Author} class represents an interesting opportunity to formulate fundamentally new classification tasks, such as writer identification from handwriting samples, but this is beyond the scope of this work.

\section{Methodology}
\label{sec:methodology}

This paper introduces the EMNIST datasets and then applies an OPIUM-based classifier to the classification tasks based on these datasets. The purpose of the classifiers is to provide a means of validating and characterizing the datasets whilst also providing benchmark classification results. The nature and organization of the EMNIST datasets are explained through the use of these classification results.

To maximize the reproducibility and accessibility of this dataset, this section carefully outlines the steps used in converting the original NIST images into a format that is directly compatible with the images from the original MNIST dataset. The conversion process used sought to reproduce the steps used in creating the original MNIST dataset (which was also created from the NIST digits) as outlined in \cite{LeCun1998}, and then to apply the same processing steps on the entire contents of the NIST Special Database 19. This conversion process, and the modifications implemented to better convert the letters in the dataset, is described in Section~\ref{sec:dataset-conversion}. 

The classifiers used to create the initial benchmark results on these new datasets are introduced in Section~\ref{sec:methodology-classifiers}. These classifiers were applied to the classification tasks defined by the EMNIST datasets which contain different assortments of letters, digits, and combinations of both. The OPIUM-based classifiers were chosen for this work as they are based on a pseudo-inverse network solution and therefore provide a deterministic, single-step analytical solution. 

\subsection{Conversion Process}
\label{sec:dataset-conversion}

The NIST Special Database 19 was originally released in 1995 and made use of an encoding and compression method based on the CCITT Group 4 algorithm \cite{Urban1992}, and packed the compressed images into a proprietary file format. Although the initial release of the database includes code to extract and process the dataset, it remains difficult to compile and run these utilities on modern systems.

In direct response to this problem, a second edition of the dataset was published in September 2016 and contains the same data encoded using the PNG file format. The work presented in this paper makes use of the original dataset and includes code and instructions to extract and convert those files. The post-processing techniques used to create the down-sampled $28 \times 28$ pixel images are directly compatible with the second edition of the dataset.

\begin{figure*}
  \centering
  \includegraphics[width=\textwidth]{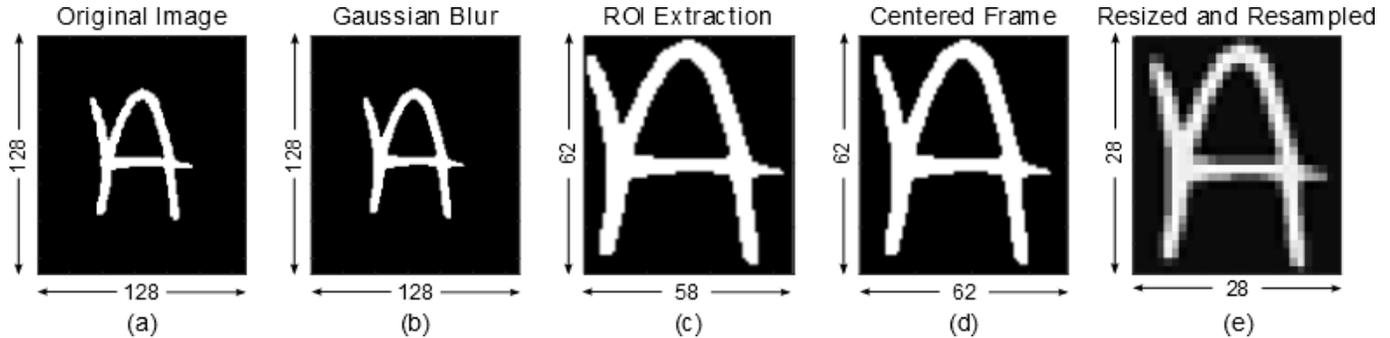}
  \caption{\textbf{Diagram of the conversion process used to convert the NIST dataset.} The original images are stored as $128 \times 128$ pixel binary images as shown in (a). A Gaussian filter with $\sigma = 1$ is applied to the image to soften the edges as shown in (b). As the characters do not fill the entire image, the region around the actual digit is extracted (c). The digit is then placed and centered into a square image (d) with the aspect ratio preserved. The region of interest is padded with a 2 pixel border when placed into the square image, matching the clear border around all the digits in the MNIST dataset. Finally, the image is down-sampled to $28 \times 28$ pixels using bi-cubic interpolation. The range of intensity values are then scaled to $[0, 255]$, resulting in the $28 \times 28$ pixel gray-scale images shown in (e).}
  \label{fig:conversion-process}
\end{figure*}

The conversion process transforms the $128 \times 128$ pixel binary images found in the NIST dataset to $28 \times 28$ pixel images with an 8-bit gray-scale resolution that match the characteristics of the digits in the MNIST dataset. An overview of the conversion process is presented in Figure~\ref{fig:conversion-process}. As described in Section~\ref{sec:the-datasets}, the NIST dataset contains $814,255$ images in four different hierarchies which affect the labeling and organization of the data. 

For this work, only the \textit{By\_Class} and \textit{By\_Merge} hierarchies are used. 
The conversion process for both is identical and only the class labels (and number of class labels) changes.

The conversion methodology follows the same overall paradigm as the conversion process used for the MNIST dataset and outlined in \cite{LeCun1998}, but makes use of a different down-sampling method to better handle the variations in shape and size of the characters in the NIST dataset.

In order to convert the dataset, each digit is loaded individually and blurred using a Gaussian filter. A bounding box is fitted to the character in the image and extracted. As the size and shape of the characters and digits vary both from class-to-class and from writer-to-writer, there is significant variance in the size of the region of interest. Whereas the original MNIST conversion technique down-sampled the digits to either a $20 \times 20$ pixel or a $32 \times 32$ pixel frame before placing it into the final $28 \times 28$ pixel frame, the technique used in this paper attempts to make use of the maximum amount of space available. 

To perform the conversion, the extracted region of interest is centered in a square frame with lengths equal to the largest dimension, with the aspect ratio of the extracted region of interest preserved. This square frame is then padded with an empty 2 pixel border to prevent the digits and characters from touching the border. Finally, the image is down-sampled to $28 \times 28$ pixels using a bi-cubic interpolation algorithm, resulting in a spectrum of intensities which are then scaled to the 8-bit range. 

\subsection{Training and Testing Splits}
\label{sec:methodology-splits}

The handwriting data used in the Special Database 19 was collected from both Census employees and high-school students. The specifications provided alongside the dataset include a suggestion that the handwritten digits from the student corpus be used as the testing set. Although the argument that the student handwriting represents a harder task and therefore should be used as an unseen testing set has merit, it does raise questions as to whether there is enough similarity and consistency between the two sets of participants. 

For this reason, the original MNIST dataset uses a different training and testing split from the one specified and recommended in the user guide supplied alongside both dataset releases. The creation of the EMNIST dataset therefore follows the same methodology used in the original MNIST paper \cite{LeCun1998} in which the original training and testing sets were combined and a new random training and testing set were drawn. The resulting training and testing datasets thereby contain samples from both the high-school students and the census employees.

\subsection{The EMNIST Datasets}
\label{sec:dataset-characteristics}

The NIST Special Database 19 contains two arrangements of segmented handwritten characters that are well suited to creating a new classification task similar to that of MNIST. These are the By\_Class and By\_Merge hierarchies introduced in Section~\ref{sec:the-datasets} and both contain the same image data but have different class labels. The underlying images were all converted to $28 \times 28$ pixel representations using the method described in Section~\ref{sec:dataset-conversion}.

These two datasets represent the full complement of NIST handwritten characters, but have an uneven number of samples per class. Due to the nature of the collection process, there are far more digit samples than letter samples, and the number of samples per class in the letter portion of the dataset is approximately equal to their frequency in the English language. As a result, four additional subsets of these dataset were produced to specifically address these issues.

\begin{figure*}
  \centering
  \includegraphics[width=\textwidth]{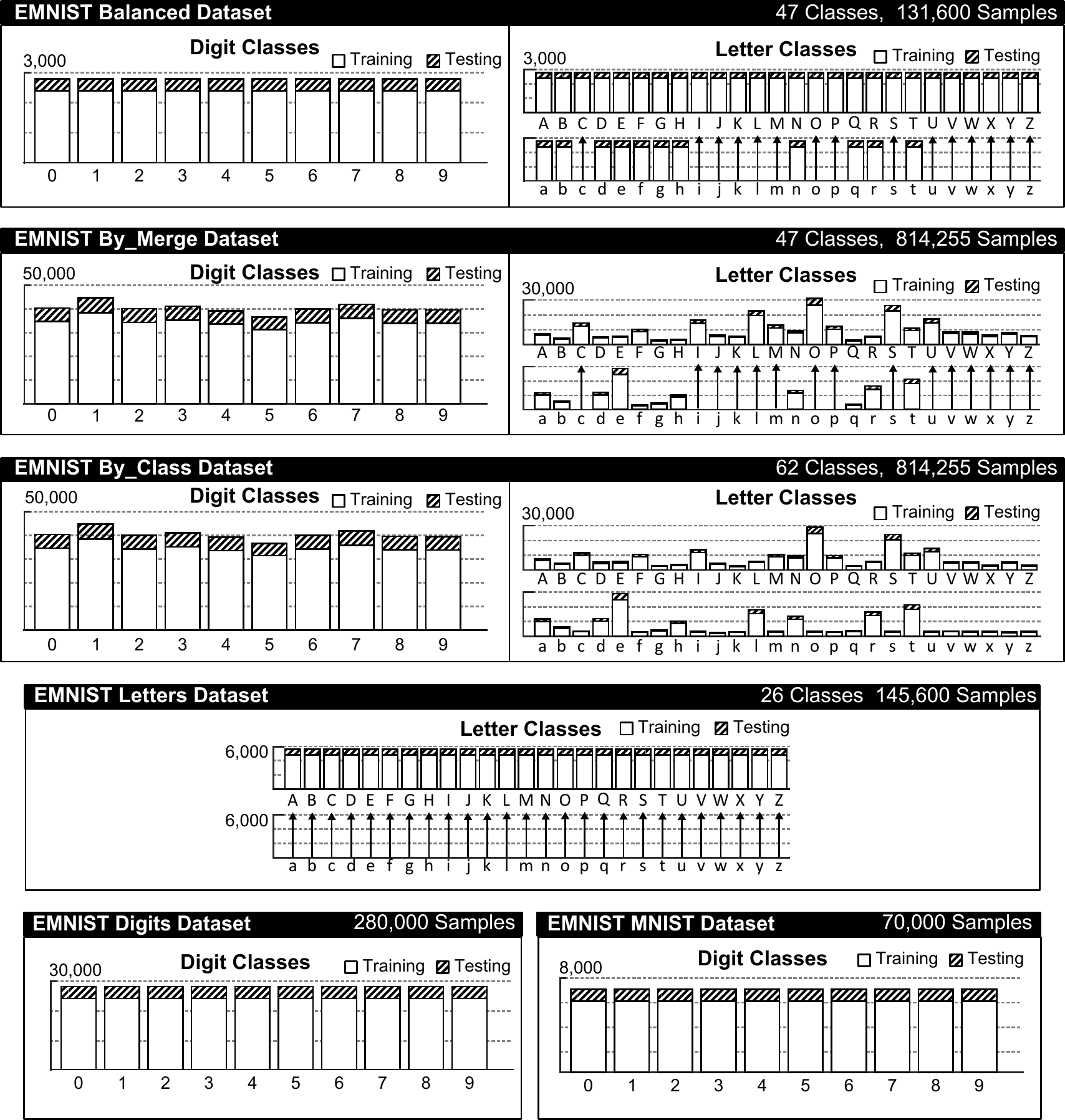}
  \caption{\textbf{Visual breakdown of the EMNIST datasets.} The class breakdown, structure and splits of the various datasets in the EMNIST dataset are shown. Each dataset contains handwritten digits, handwritten letters or a combination of both. The number of samples in each class is shown for each dataset and highlights the large variation in the number of samples in the unbalanced datasets. The training and testing split for each class is also shown using either solid or hatched portions of the bar graphs. In the datasets that contain merged classes, a vertical arrow is used to denote the class into which the lowercase letter is merged. }
  \label{fig:dataset-breakdown}
\end{figure*}

Figure~\ref{fig:dataset-breakdown} provides a summary of the contents of the six datasets that comprise the EMNIST datasets. The figure shows the included classes and number of samples per class in each of the six datasets. Additionally, it shows the training and testing splits for each dataset. 

The EMNIST By\_Class and EMNIST By\_Merge datasets both contain the full 814,255 characters and differ only in the number of assigned classes. Thus, the distribution of samples in the letter classes varies between the two datasets. The number of samples in the digits classes remains unchanged between the two datasets.

The EMNIST Balanced dataset is intended to be the most widely applicable dataset as it contains a balanced subset of all the By\_Merge classes. The 47-class dataset was chosen over the By\_Class dataset to avoid classification errors resulting purely from misclassification between uppercase and lowercase letters. 

The EMNIST Letters dataset seeks to further reduce the errors occurring from case confusion by merging all the uppercase and lowercase classes to form a balanced 26-class classification task. In a similar vein, the EMNIST Digits class contains a balanced subset of the digits dataset containing 28,000 samples of each digit. 

Finally, the EMNIST MNIST dataset is intended to exactly match the size and specifications of the original MNIST dataset. It is intended to be a drop-in replacement for the original MNIST dataset containing digits created through the conversion process outlined in Section~\ref{sec:dataset-conversion}. It is used primarily to validate and characterize the conversion process against the original MNIST dataset.

\subsection{Validation Partitions}
\label{sec:methodology-validation-sets}

Many iterative training algorithms make use of a validation partition to assess the current performance of a network during training. This needs to be separate from the unseen testing set to maintain the integrity of the results. Instead of including a separate validation set for each class, the balanced datasets in the EMNIST dataset contain a specially balanced subset of the training set intended specifically to be used for validation tasks. 

\begin{table}[]
\centering
\caption{Structure and Organization of the EMNIST datasets.}
\label{tab:emnist-validation-composition}
\begin{tabular}{@{}llcccc@{}}
\toprule
Name & Classes & No. Training & No. Testing & Validation & Total \\ \midrule
By\_Class & 62 & 697,932 & 116,323 & No & 814,255 \\
By\_Merge & 47 & 697,932 & 116,323 & No & 814,255 \\
& & & & & \\
Balanced & 47 & 112,800 & 18,800 & Yes & 131,600 \\ 
Digits & 10 & 240,000 & 40,000 & Yes & 280,000 \\ 
Letters & 37 & 88,800 & 14,800 & Yes & 103,600 \\ 
MNIST & 10 & 60,000 & 10,000 & Yes & 70,000 \\
\bottomrule
\end{tabular}
\end{table}

Table~\ref{tab:emnist-validation-composition} contains a summary of the EMNIST datasets and indicates which classes contain a validation subset in the training set. In these datasets, the last portion of the training set, equal in size to the testing set, is set aside as a validation set. Additionally, this subset is also balanced such that it contains an equal number of samples for each task. If the validation set is not to be used, then the training set can be used as one contiguous set. 

\subsection{Classifiers}
\label{sec:methodology-classifiers}

The classification results provided in this work are intended to form a benchmark for the datasets provided. A simple three-layer ELM network, as described in \cite{Huang2006}, was used to perform the classification and it is expected that deeper and more sophisticated networks will provide better classification performance. In addition, a linear classifier was also trained on the dataset. This classifier represents the analytical pseudo-inverse solution for a network without a hidden layer.

The pseudo-inverse required for the ELM cannot be calculated in a single step due to size of the dataset and the networks are instead trained using the Online Pseudo-Inverse Update Method (OPIUM) \cite{VanSchaik2015b}. This method iteratively calculates the exact pseudo-inverse solution for the output weights and allows the network to handle datasets of any size. 

The OPIUM-based classification networks were tested over a range of hidden layer sizes. A random training order was used and kept constant throughout all the tests. Multiple trials with different random input weights and the same hidden layer size were conducted and the mean classification accuracy and standard deviation reported when appropriate. The linear classifier, although trained using the same iterative pseudo-inverse technique, does not contain a hidden layer and therefore only a single result per dataset was obtained.

The purpose of this paper is to present and characterize the datasets. The OPIUM-based training methods were selected as they generate an analytical solution and does not require multiple iterations over the dataset. The outcomes of these networks are deterministic, given the same network structure and training order. The results in this paper do not represent cutting edge techniques, but rather serve as an instructive baseline and a means of exploring and validating the datasets.

% Linear classification results
% Balanced: 50.93\%
% By\_Merge: 50.51\%
% By\_Class: 51.80\%
% MNIST: 85.11\%
% Letters:55.78\%
% Digits: 84.70\%

\section{Results}
\label{sec:results}

This section presents the results for the various classification tasks based on the EMNIST datasets. The classifiers used are all based on the OPIUM classifiers described in Section~\ref{sec:methodology-classifiers} and the overarching methodology aimed to keep the structure and the network parameters constant between experiments and between trials. Only the hidden layer size and the random weights were altered from network to network. The results section is broken down into different sections for each type of dataset and provides a summary and discussion of the results. 

\subsection{EMNIST Balanced Dataset}
\label{sec:results-balanced}

The EMNIST Balanced dataset is intended to provide the most fair and consistent classification task derived from the NIST Special Database 19. It contains a subset of the By\_Merge dataset containing an equal number of samples for each class. The intention of this dataset is to provide a classification task that is fair, balanced and sufficiently challenging. 

\begin{figure}
  \centering
  \includegraphics[width=0.45\textwidth]{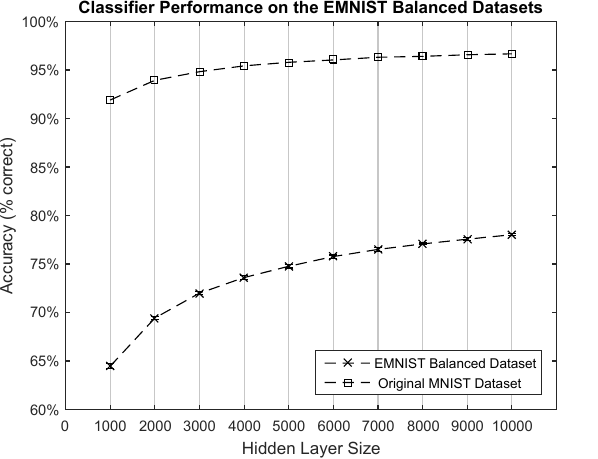}
  \caption{\textbf{Comparison of performance on the EMNIST Balanced Dataset and the original MNIST dataset.} The classification results for OPIUM networks of varying hidden layer size trained on the EMNIST Balanced dataset. Twenty trials of each experiment were conducted and the mean accuracy and standard deviation plotted above. Additionally, the classification accuracy achieved using the same network on the original MNIST dataset is also show on the same axes.}
  \label{fig:results-emnist-balanced}
\end{figure}

Figure~\ref{fig:results-emnist-balanced} presents the classification results of the OPIUM-based classifiers of varying hidden layer size trained on the entire training set of the EMNIST dataset. For the purposes of these experiments, the validation portion of the training set was included in the training, with all results presented representing the final accuracy achieved on the unseen testing set. The accuracy of the classification task increased with each hidden layer size, reaching a maximum mean accuracy of $78.02\% \pm 0.09\%$ over twenty trials of networks containing 10,000 hidden layer neurons. Larger hidden layer sizes could not be explored due to memory and processing constraints. By contrast, the linear classifier trained on this dataset achieved a classification accuracy of $50.93\%$.

Figure~\ref{fig:results-emnist-balanced} also includes the classification results achieved when using networks with identical structure to classify the digits from the original MNIST dataset. These networks made use of the same hidden layer size and the same random weights for each trial, and the results are consistent with previously published results for these networks \cite{VanSchaik2015b}. It is clear from the figure that the EMNIST dataset succeeds in providing a more challenging classification task for these networks, whilst still maintaining the structure, simplicity and accessibility of the original MNIST dataset.

\subsection{EMNIST By\_Merge and By\_Class Dataset Results}
\label{sec:results-bymerge-byclass}

The EMNIST By\_Merge an EMNIST By\_Class datasets both contain the full 814,255 characters contained in the original NIST Special Database 19. The primary differences between the two datasets are the number of classes, the number of samples per class, and the order in which the characters appear in the dataset. The By\_Class dataset contains the full 62 classes comprising 10 digit classes, 26 lowercase letter classes, and 26 uppercase letter classes. The By\_Merge dataset merges certain uppercase and lowercase characters, containing 10 digit classes and 47 letter classes. The number of characters per class varies dramatically throughout both datasets, as is clearly visible in Figure~\ref{fig:dataset-breakdown}. The datasets also contain vastly more digit samples than letter samples. 

\begin{figure}
  \centering
  \includegraphics[width=0.45\textwidth]{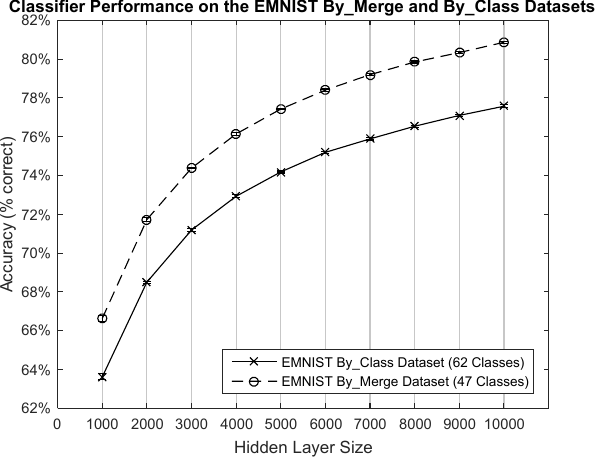}
  \caption{\textbf{Classifier performance on the EMNIST By\_Class and By\_Merge Datasets.} The classification results for OPIUM-based classifiers of varying sizes applied to the two full EMNIST datasets are shown in the above figure. Ten trial of each experiment were performed. Both datasets contain all 814,255 training, with the By\_Class dataset containing the full 62 classes and the By\_Merge dataset merging certain letter classes to result in 47 classes. The results show that the merged 47-way classification task with the By\_Merge dataset outperforms the By\_Class dataset at all hidden layer sizes, although both suffer from significant variations in the letter classes accuracy from trial to trial which is overshadowed by the much larger set of samples in the digit classes.}
  \label{fig:results-emnist-bymergebyclass}
\end{figure}

A set of OPIUM-based classifiers were used to explore the nature of the classification tasks designed around these two datasets. The full training set was used to train the classification networks with varying hidden layer sizes, which were subsequently tested on the testing set. Ten trials of each experiment were performed and the results of the classifiers are shown in Figure~\ref{fig:results-emnist-bymergebyclass}. The graph shows the accuracy for both datasets in terms of the percentage of characters correctly identified and shows that the By\_Merge dataset outperforms the By\_Class dataset at every hidden layer size. The By\_Merge dataset achieved a peak accuracy of $80.87\% \pm 0.05\%$ over ten trials using 10,000 hidden layer neurons, whilst the By\_Class dataset achieved only $77.57\% \pm 0.08\%$ using a network of the same size. The linear classifiers achieved an accuracy of  $50.51\%$ and $51.80\%$ accuracy on the By\_Merge and By\_Class datasets respectively.

The performance difference between these two datasets is primarily due to misclassification between uppercase and lowercase letters in the By\_Class dataset and serves to further highlight the motivation for merging certain letter classes to create the By\_Merge dataset. Although the By\_Merge dataset does correct much of the case-related misclassification, there are still several letters and digits that are commonly confused by the classifiers, primarily the lowercase 'L' class, the digit '1' and the lowercase 'I' class. These appear as separate classes in the By\_Merge dataset and contribute a significant portion of the errors. It is possible that more sophisticated classifiers will be able to better distinguish between the subtle differences in these characters. 

\subsection{EMNIST Letters Results}
\label{sec:results-letters}

The EMNIST Letters dataset was created to mitigate the issues regarding case and the misclassification of letters and digits that plague the By\_Class and By\_Merge datasets. This is accomplished by combining the uppercase and lowercase versions of each letter into a single class and removing the digit classes entirely. This dataset therefore provides a different classification task from the other datasets and poses a letter-only classification task in the spirit of the original MNIST dataset. As certain letters have distinctly different uppercase and lowercase representations, the classifiers are required to associate two different representations of a letter with a single class label. 

\begin{figure}
  \centering
  \includegraphics[width=0.45\textwidth]{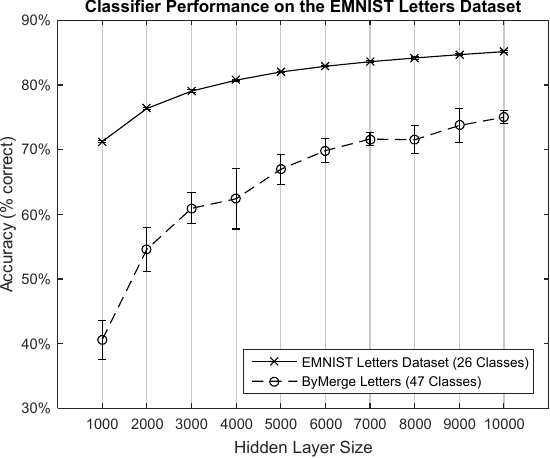}
  \caption{\textbf{Classifier performance for the EMNIST Letter Dataset.} The classification performance achieved on the EMNIST Letters dataset over a range of hidden layer sizes is presented above. The accuracy is measured in percentage of characters correctly classified and was calculated over ten independent trials of the experiment. For comparison, the performance of the same network on the letters contained in the EMNIST By\_Merge dataset is also shown. This represents the performance of the mixture of uppercase and lowercase letters found in that dataset. It is clearly visible from the above plot that the variance due to uppercase and lowercase classifications is drastically reduced with the EMNIST Letters dataset.}
  \label{fig:results-emnist-letters}
\end{figure}

In order to demonstrate the benefits of this dataset organization, OPIUM-based classifiers of varying hidden layer size were trained on the training set and then tested on the unseen testing set. Ten trials of each experiment were performed, and the results achieved with the classifier are presented in Figure~\ref{fig:results-emnist-letters}. The accuracy achieved with the network increased with the size of the hidden layer, achieving a maximum accuracy of $85.15\% \pm 0.12\%$ using 10,000 hidden layer neurons. As expected, the accuracy of this dataset does outperform the accuracy of $80.87\% \pm 0.05\%$ achieved on the full EMNIST By\_Merge dataset, although this comparison is marred by the uneven number of samples per class in the testing set of the EMNIST By\_Merge dataset.

The linear classifier trained on the EMNIST Letters dataset produced an accuracy of $55.78\%$, a moderate increase over the accuracy achieved on the By\_Merge dataset.

Figure~\ref{fig:results-emnist-letters} also includes the results of applying the same classification networks to just the letter classes contained in the EMNIST By\_Merge dataset (37 of the available 47 classes). The same number of training and testing samples were randomly extracted from the dataset and used to create the 37-class classification task containing a mixture of merged letter classes, uppercase classes and lowercase classes. Mirroring the structure of the experiments conducted on the EMNIST Letters dataset, ten trials of each experiment were conducted on the dataset. The results of these classifiers are shown on the same set of axes.

Immediately clear from the results is the lower accuracy and significantly higher variance when using the mixed case version of the dataset. This is primary due to mis-classifications between the uppercase and lowercase versions of the same letters, which varies significantly from trial to trial of the same experiment with different sets of random input-layer weights. Through checking a selection of the original handwriting forms, it appears as if this problem results from handwriting inconsistencies in individuals rather than mis-labeling in the dataset. The loss of scale in the conversion process also removes any cues that result from the relative size of the handwritten characters, although this too varies between individuals. The network achieved a best-case accuracy of $75.0\% \pm 1.03\%$ with a network containing 10,000 hidden layer neurons. It is likely that the standard deviation would decrease with additional trials, but the stated values were achieved using the same number of trials as used to obtain the results on the EMNIST Letters dataset.

\begin{figure}
  \centering
  \includegraphics[width=0.45\textwidth]{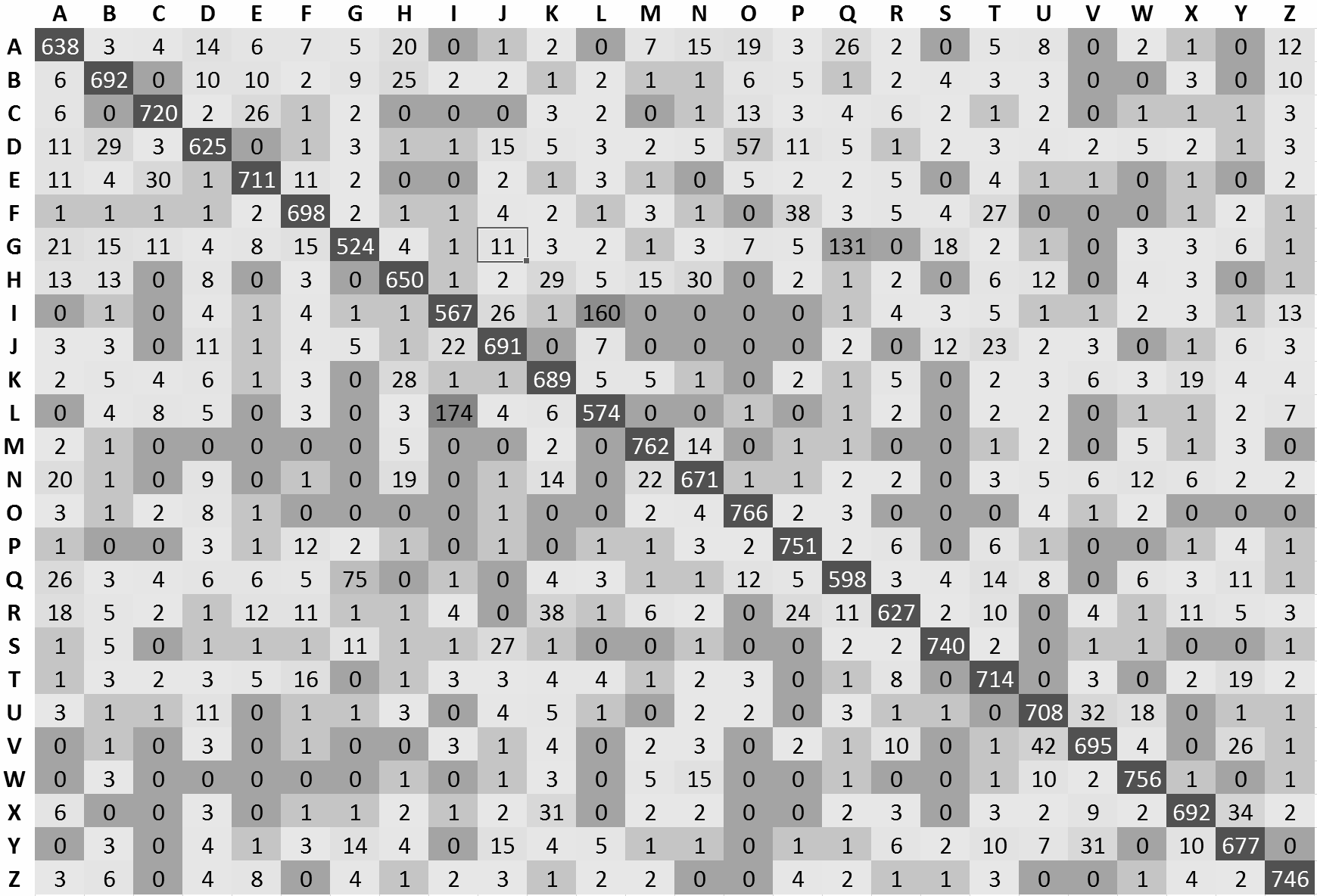}
  \caption{\textbf{Confusion matrix for the EMNIST Letters Dataset for a network with 10,000 hidden layer neurons.} The confusion matrix for a single trial of an OPIUM-based classifier containing 10,000 hidden layer neurons and trained and tested using the full EMNIST Letters dataset. The confusion matrix shows that most of the confusions occur between classes that inherently contain ambiguity, primarily between the I and L classes. The G and Q classes also appear to suffer from ambiguity resulting from the similarity in their lowercase representations. }
  \label{fig:results-emnist-letters-confusions}
\end{figure}

Figure~\ref{fig:results-emnist-letters-confusions} shows a typical confusion matrix achieved when using an OPIUM-based classifier containing 10,000 hidden layer neurons and trained using the EMNIST Letters dataset. The confusion matrix shows the letter pairs that are most commonly misclassified, with the letters I and L being the most commonly confused letter, followed by the G and Q class. These same confusions are present in the full By\_Merge dataset, further complicating the errors that result between letters of the same case. The problem is further complicated when including the digit classes as well, which is the primary motivating factor for restricting the EMNIST Letters dataset to just the 26 letter classes. In the above example, the class with the lowest classification accuracy is the G class, which achieved only $65.5\%$ accuracy. The best performing class was the O class, which achieved a classification accuracy of $95.8\%$. 

\subsection{EMNIST Digit Dataset Results}
\label{sec:results-emnist-digits}

Complementing the EMNIST Letters dataset is the EMNIST Digits dataset, which contains only the digit classes from the EMNIST By\_Class dataset. It contains the largest possible subset of digits containing an equal number of samples for each digit class. It is intended to serve as a direct drop-in replacement for the original MNIST dataset as it maintains the input size, number of labels and the structure of the training and testing splits.

In addition to the EMNIST Digits dataset, the EMNIST MNIST dataset contains the same number of digits as the original MNIST dataset. It was intended to serve as a means of verifying the efficacy of the conversion process and allow for a direct comparison to the original MNIST dataset. This dataset also contains a balanced number of each class and matches the training and testing split of the original dataset.

\begin{figure}
  \centering
  \includegraphics[width=0.45\textwidth]{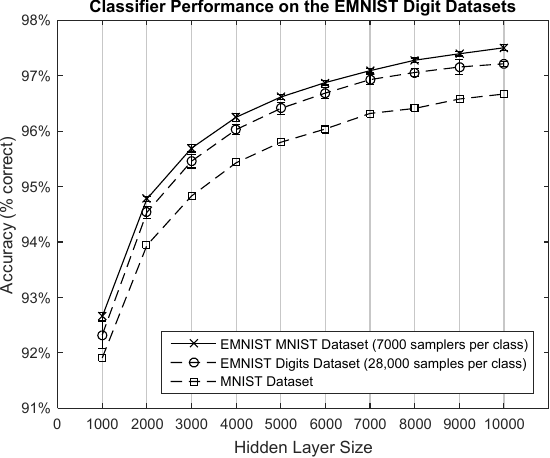}
  \caption{\textbf{Classifier performance on the EMNIST Digits and EMNIST MNIST datasets.} The classification performance for the classifiers trained on the two digit datasets are shown in the above figure. Ten trials of each experiment were performed. }
  \label{fig:results-emnist-digits-performance}
\end{figure}

Figure~\ref{fig:results-emnist-digits-performance} presents the results of  OPIUM-based classifiers of varying hidden layer sizes trained on both the EMNIST Digits and the EMNIST MNIST datasets. Additionally, the same networks were used on the original MNIST dataset and the results included as a point of reference and as a means of validating and characterizing the conversion process used in this work. Ten trials of each network were performed, with the exact same networks being used for all three datasets. The EMNIST MNIST dataset achieved a maximum accuracy of $97.50\% \pm 0.05\%$ with a 10,000 hidden layer network, slightly outperforming the EMNIST Digits dataset which achieved $97.22\% \pm 0.05\%$.

The linear classifiers produced similar results, with the EMNIST Digits dataset producing an accuracy of $84.70\%$ and EMNIST MNIST producing a slightly better $85.11\%$. These results are much higher than those achieved with the EMNIST Letters dataset, indicating that the letters constitute a task that is less linearly separable. This is directly comparable with the linear classification result of $88\%$ achieved on MNIST without preprocessing \cite{LeCun1998}.

The difference between the performance of the two datasets is relatively small, and is most likely due to the increased variability in the full 280,000 digits in the EMNIST Digits dataset. It appears as if the classification of handwritten digits remains an easy task, even with the vast increase in the number of samples. It is also immediately evident from Figure~\ref{fig:results-emnist-digits-performance} that the performance of both the EMNIST Digits and the EMNIST MNIST datasets outperform the original MNIST dataset at every network size. As identical networks were used to obtain all three results, it demonstrates that the conversion process used in this work results in a more separable problem than the original MNIST dataset. This is most likely due to the conversion process used in this work ensuring that the character fills as much of the $28 \times 28$ pixel frame as possible. This serves to reduce the number of blank pixels around each digit and therefore allows for more information to be captured in the process. 

The EMNIST MNIST dataset also contains a slightly different set of digits from the original MNIST dataset, primarily due to the need to balance the number of samples in each class. Although this may have impacted the accuracy achieved on the dataset somewhat, it does not explain the significant and consistent improvement in accuracy achieved using the EMNIST digits.
 
\section{Discussion}
\label{sec:results-summary}

Table~\ref{table-results-summary} provides a summary of the classification accuracies achieved on the EMNIST datasets. The accuracy of each classification task increased with the size of the hidden layer in the network, resulting in a best performance with network sizes of 10,000 hidden layer neurons. Due to memory and processing constraints, it was not possible to explore larger networks. 

% \begin{table}[]
% \centering
% \caption{Summary of results for the OPIUM-based classification networks of 10,000 hidden layer neurons (n = 10)}
% \label{table-results-summary}
% \begin{tabular}{llll}
% \toprule
% Dataset          &  & Mean    & STD    \\ \midrule \\
% EMNIST Balanced  &  & 78.02\% & 0.09\% \\
% EMNIST By\_Merge &  & 80.87\% & 0.05\% \\
% EMNIST By\_Class &  & 77.57\% & 0.08\% \\
% EMNIST Letters   &  & 85.15\% & 0.12\% \\
% EMNIST Digits    &  & 97.22\% & 0.05\% \\
% EMNIST MNIST     &  & 97.50\% & 0.05\% \\
% \bottomrule
% \end{tabular}
% \end{table}

\begin{table}[]
\centering
\caption{Summary of the results for the linear and OPIUM-based classifiers on the EMNIST dataset.}
\label{table-results-summary}
\begin{tabular}{llll}
\toprule
             &  & Linear Classifier & OPIUM Classifier \\ \midrule
Balanced     &  & $50.93\%$ & $78.02\% \pm 0.92\%$     \\
By Merge     &  & $50.51\%$ & $72.57\% \pm 1.18\%$     \\
By Class     &  & $51.80\%$ & $69.71\% \pm 1.47\%$     \\
Letters      &  & $55.78\%$ & $85.15\% \pm 0.12\%$     \\
EMNIST MNIST &  & $85.11\%$ & $96.22\% \pm 0.14\%$     \\
Digits       &  & $84.70\%$ & $95.90\% \pm 0.40\%$     \\
\bottomrule
\end{tabular}
\end{table}

More sophisticated methods, such as those incorporating multiple layers, are expected to produce networks capable of achieving better performance. As these are the first results on this dataset, there are no directly comparable results, but these results are consistent with prior work accomplished using the NIST Special Database 19.

The results most relevant to this work are those of Cire\c{c}an et al., who made use of the NIST Special Database 19 to further validate their convolutional neural network \cite{Ciresan2011}. Using a committee of seven deep networks, they achieved $0.27\%$ error on the MNIST dataset. They then made use of the same network structure for the NIST task and thereby performed a similar conversion process to the one outlined in this work to convert the NIST images into the same input size as the original MNIST data. 

Their approach differs slightly from the one used in this paper in that they resized the digits to a $20 \times 20$ bounding box and then centered this inside a $29 \times 29$ pixel frame. They also did not release either the code or the dataset. 

Their results, which are based on a similar dataset, provide an estimation of the expected performance when using more sophisticated classification techniques. They achieved an equivalent of $88.12\% \pm 0.09\%$ on the By\_Class dataset and $91.79\% \pm 0.11\%$ on the By\_Merge dataset. 

They also reported an accuracy of $92.42\% \pm 0.09\%$ on the letters portion of the dataset and $99.19\% \pm 0.02\%$ on the digits portion, although these are not directly comparable as they made use of the full complement of available digits rather than the balanced sets found in the EMNIST dataset.

Another work of interest is that of Milgram et al., who made use of the digits in the NIST dataset as a classification task for comparing post-processing methods for multi-class probabilities with Support Vector Machines \cite{Milgram2005}. Their work makes use of a different set of training and testing splits, and achieved an accuracy of $98.11\%$ on the digit classification task. 

Granger et al. also made use of the full complement of digits in the NIST dataset to test their fuzzy ARTMAP neural network architecture \cite{Granger2007} and achieved an accuracy of $96.29\%$. 

Given that the classifiers presented in this work were chosen for their simplicity rather than their cutting-edge performance, these comparative results serve to further validate the datasets. Although not always directly comparable, the results from all prior work on the NIST Special Database 19 dataset are in-line with the results presented above and give a strong indication of the potential performance achievable using the EMNIST datasets.

\section{Conclusions}
\label{sec:conclusions}

This paper introduced the EMNIST datasets, a suite of six datasets intended to provide a more challenging alternative to the MNIST dataset. The characters of the NIST Special Database 19 were converted to a format that matches that of the MNIST dataset, making it immediately compatible with any network capable of working with the original MNIST dataset. Benchmark results are provided which are consistent with previously published work using the NIST Special Database 19. Additionally, a comparison of the performance of the classification task on a subset of digits against the original MNIST dataset served to further validate the conversion process.

The EMNIST datasets therefore provide a new classification benchmark that contains more image samples, more output classes, a more varied classification task, and a more challenging classification task than MNIST whilst maintaining its structure and nature. This therefore represents a new and modern performance benchmark for the current generation of classification  and learning systems.

\section*{Acknowledgment}
\label{sec:acknowledgments}
The authors would like to thank the organizers and funders of the Capo Caccia and Telluride Cognitive Neuromorphic Engineering Workshops, at which the initial steps of this work were completed.

\bibliographystyle{IEEEtran}
\bibliography{Papers-NISTPaper}

\end{document}